\documentclass{article}

\usepackage{arxiv}

\usepackage[utf8]{inputenc} 
\usepackage[T1]{fontenc}    
\usepackage{hyperref}       
\usepackage{url}            
\usepackage{booktabs}       
\usepackage{amsfonts}       
\usepackage{nicefrac}       
\usepackage{microtype}      
\usepackage{lipsum}
\usepackage{graphicx}
\graphicspath{ {./images/} }

\title{A novel Multi to Single Module for small object detection}

\author{
 Xiaohui Guo \\
  School of Guangdong Ocean University\\
  \texttt{guoxiaohui1@stu.gdou.edu.cn} \\
   \And
 Haitao Liu \\
  School of Guangdong Ocean University\\
  \texttt{guoxiaohui1@stu.gdou.edu.cn} \\
  \And
 Weng \\
  School of Guangdong Ocean University\\
  \texttt{guoxiaohui1@stu.gdou.edu.cn} \\
}

\begin{document}
\maketitle
\begin{abstract}
Small object detection presents a significant challenge in computer vision and object detection. The performance of small object detectors is often compromised by a lack of pixels and less significant features. This issue stems from information misalignment caused by variations in feature scale and information loss during feature processing. In response to this challenge, this paper proposes a novel the Multi to Single Module (M2S), which enhances a specific layer through improving feature extraction and refining features. Specifically, M2S includes the proposed Cross-scale Aggregation Module (CAM) and explored Dual Relationship Module (DRM) to improve information extraction capabilities and feature refinement effects. Moreover, this paper enhances the accuracy of small object detection by utilizing M2S to generate an additional detection head. The effectiveness of the proposed method is evaluated on two datasets, VisDrone2021-DET and SeaDronesSeeV2. The experimental results demonstrate its improved performance compared with existing methods. Compared to the baseline model (YOLOv5s), M2S improves the accuracy by about 1.1\% on the VisDrone2021-DET testing dataset and 15.68\% on the SeaDronesSeeV2 validation set.
\end{abstract}


\section{Introduction}
Small object detection (\cite{Zhou2021IntelligentSO},\cite{Nguyen2020AnEO},\cite{Huang2022SmallOD},\cite{Yu2022FastCD},\cite{Liu2022SFYOLOv5AL}) has long been a challenge in object detection, which aims to accurately detect small objects (those under 32 pixels by 32 pixels) with very few visual features in the image. A detection model consisting of a backbone network, an encoder, and a decoder is proposed in You Only Look One-level Feature (YOLOF) \cite{Chen2021YouOL}. It is proposed in YOLOF that by choosing the right scale features for output to a particular level with multiple-input and single-output, comparable performance to multiple-input and multiple-output can be obtained. In the small object detection task, this paper proposes a multiple-input to the single-output module that outputs with a low-level feature with high resolution. It provides the detector with more accurate and richer information, thereby improving its performance.
\par

\begin{figure}[t]
    \centering
    \includegraphics[width=12cm]{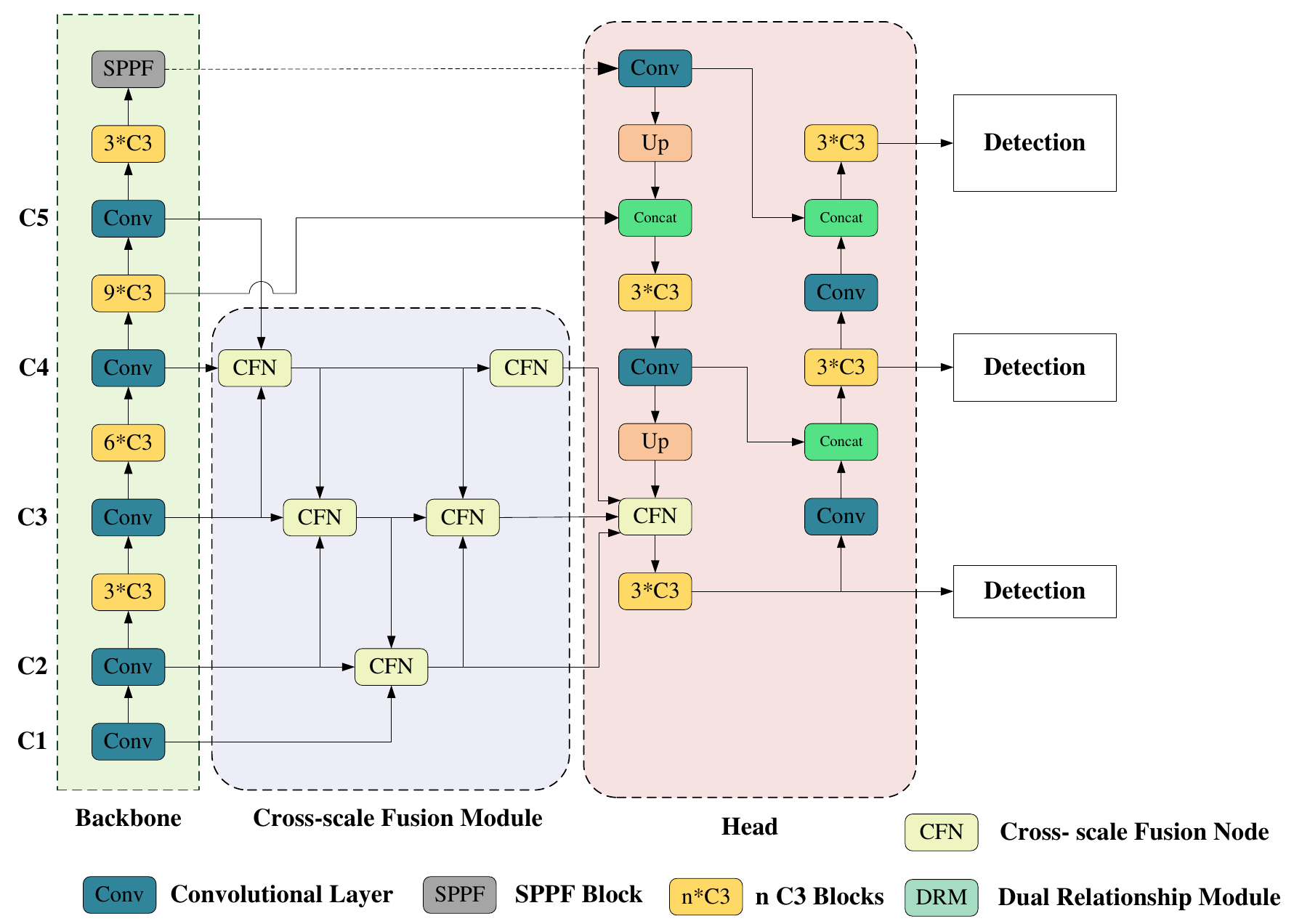}
    \caption{The architecture of M2S. a) Darknet53 backbone with the CAM block at the start. b) The CAM contains multiple cross-scale fusion nodes. c) The DRM module is inserted afterwards  the CRM and is represented by a pink background with rounded rectangles. d) Cn,\ $n\in\left\{{1,2,3,4,5}\right\} $   denotes the different layer features in the backbone network.}
    \label{fig:1}
\end{figure}

Feature Pyramid network (FPN) \cite{Lin2016FeaturePN}  success is due to the idea of divide and conquer. Divide and conquer ensures that image features are available in different layers, whereas if only a unidirectional flow of information is delivered, then FPN ends up with a limitation at each layer. The result is that each layer of character can only be thinking within its own perspective. Thus, previous work has been devoted to increasing the interaction of different levels of feature and allowing more global thinking at each level. FPN appears as the backbone network of excellent detectors SSD \cite{Liu2015SSDSS}, YOLO \cite{Redmon2015YouOL}, and RCNN \cite{Mnih2014RecurrentMO}. EffientDet \cite{Tan2019EfficientDetSA}uses a cross-edge connecting BiFPN and repeats it to obtain better accuracy and efficiency. Although the previous works have shown gratifying performance, object detection still has a problem: the poor detection of small-scale instance objects. Based on this problem, this paper proposes a module to enhance communication at different network levels and improve feature extraction efficiency. The Cross-scale Fusion Module aims to better aggregate more features and improve the feature extraction capability of the model, which is shown in Fig.\ref{fig:1}.

\par 
For attention, there is significant efficiency in the usage of contextual information about features. However, the potential relationships between multi-level feature are difficult to exploit fully using one type of attention alone. Inspired by CBAM \cite{Woo2018CBAMCB} and Jiont-attention \cite{Ma2022JointattentionFF}, the attention mechanism is oriented toward the channel and spatial dimensions. An innovative Dual Relationship Module (DRM) combining spatial attention and channel attention mechanisms has been introduced. The information obtained by CBAM can only be derived from a single feature of the input resulting in a lack of rich information. The paper presents a multi-input to single-output attention module for information where different layers of features have different preferences. Therefore, DRM not only combines two attention mechanisms but also exploits the characteristics of three-level feature information to achieve more effective semantic information enhancement and supplementation. The DRM is embedded subsequent to the CAM and its overview diagram is shown in Fig.\ref{fig:1}.
\par
In the validation phase, we validate our method on two datasets with a large proportion of small object datasets, VisDrone2021-DET \cite{Cao2021VisDroneDET2021TV} and SeaDronesSeeV2 \cite{Varga2021SeaDronesSeeAM}. Each of these two datasets has a large number of small-scale objects and is an excellent way to evaluate the effectiveness of our approach. In addition, ablation experiments were incorporated into the experiment to verify the role of the individual modules. Furthermore, a comparison experiment has been built to compare the performance of the method in this paper with that of the previous attention modules.
\par In summary, the overall performance of the detection model is improved by the above two innovations. The contributions of this work are summarized as follows:
\begin{itemize}              \item 		A Cross-scale Aggregation Module (CAM) for fusing five-level feature of a backbone network and fusing them into three-level feature is proposed. In this paper, CAM is embedded between the backbone and neck networks to improve the perspective of a single layer of features. 
\item	Combining spatial attention and channel attention this paper proposes a Dual Relationship Module (DRM) that compensates for the shortcomings of a single-dimensional attention mechanism. Meanwhile, using DRM, the three-level feature from Cross-scale Fusion Module is also aggregated into a one-level feature.
\item We evaluate our proposed small object detection approach on two public benchmark datasets, VisDrone2021-DET and SeaDronesSeeV2. And compared several state-of-the-art methods. The experimental results demonstrate the superior performance of our method for small object detection.
\end{itemize}

\section{Related Work}
\subsection{Object Detetion}
There are two types of current object detectors, one-stage with sliding-window and two-stage with region proposals. The latter includes two-stage for object detection: generation of regional proposals, classification, and modification of position such as RCNN, Fast RCNN \cite{Girshick2015FastR}, Faster RCNN \cite{Ren2015FasterRT}, and Mask RCNN \cite{He2017MaskR}. Contrary to the two-stage object detectors, the one-stage object detectors perform the regression and classification tasks directly, without the region of interest generation step. Consequently, the one-stage detectors have more efficient detection and require less computer performance. 
\par OverFeat \cite{Sermanet2013OverFeatIR}was the first application of a CNN-based single-stage target detector. YOLO \cite{Redmon2015YouOL} and SSD \cite{Liu2015SSDSS} were introduced after it, and SSD achieved great performance.YOLO and YOLOv2 \cite{Redmon2016YOLO9000BF} only use the last output feature of the backbone. They can both achieve a very fast detection speed, but at the expense of accuracy. YOLOv3 \cite{Redmon2018YOLOv3AI} is based on YoLov1, with an FPN architecture and a backbone that was changed from DarkNet-19 to DarkNet-53. YOLOv4 achieves superior performance with these improvements and modifications.YOLOv5 adds data enhancement and FOCUs modules to the v3 backbone and replaces IOU with GIOU \cite{Rezatofighi2019GeneralizedIO} and NMS with DiouNMS respectively.  Our improved network structure will be compared with YOLOv5 with backbone network of CSPDarkNet53 \cite{Wang2019CSPNetAN}. The FPN works as the basis for a detector model that is also used in a variety of popular models, such as RetinaNet \cite{Lin2017FocalLF}, FCOS \cite{Tian2019FCOSFC}, and their variants.
\subsection{Multi-scale Feature Operations}
Multi-scale feature \cite{Huang2019MaskSR, Wang2019PANetFI, hpgn, Ghiasi2019NASFPNLS, hsgm} is highly significant in the effect of mapping from the input image to the regression and classification workflow. Earlier detectors generally extracted the input directly, with the backbone network's data stream simply flowing directly to the head network and completing the prediction task. Mask Scoring R-CNN (MS-RCNN) \cite{Huang2019MaskSR} and SSD to perform regression and prediction tasks are based on feature pyramids, by picking the corresponding scale feature output from the feature pyramid. In an innovative work, FPN proposes an augmentation feature architecture that effectively improves detector performance by adding a top-down path to fuse multi-scale features. Inspired by previous work, Prototype Alignment Network (PANet) \cite{Wang2019PANetFI} adds a bottom-up path enhancement to the FPN. Based on the idea of multiple fusion enhancement and skip connections, the bi-directional FPN proposes efficient bidirectional cross-scale connections and weighted feature fusion after being inspired by EfficientDet \cite{Tan2019EfficientDetSA}. Efficient Multi-Resolution Network (EMRN) \cite{emrn} proposes a multi-resolution features dimension uniform module to fix dimensional features from images of varying resolutions. Furthermore, M2det \cite{Zhao2018M2DetAS} proposes a U-shaped connection structure. NAS-FPN \cite{Ghiasi2019NASFPNLS} achieves satisfactory results by searching a pre-defined topology of the feature network, but requires better computational performance from the GPU. 
\subsection{Attention Mechanisms}
Attention mechanisms \cite{Mnih2014RecurrentMO, shen2023triplet, Jaderberg2015SpatialTN, camnet, Woo2018CBAMCB} enable the selection of information that is required to obtain a target, with outputs dynamically weighted according to input characteristics. Demonstrates efficient, attention mechanism is added to the object detectors. RAM \cite{Mnih2014RecurrentMO} is a pioneer in introducing attention mechanisms into deep neural networks. RAM showed that the attention mechanisms are effective in working with neural networks, then related works have been presented sequentially. STN \cite{Jaderberg2015SpatialTN} introduces a sub-network to select important regions. SENet \cite{Hu2017SqueezeandExcitationN} differs from RAM and STN in the way that SENet innovatively introduces attention mechanisms to the detector to refine the channels of features. Related work has been launched with CBAM \cite{Woo2018CBAMCB}, ECANet \cite{Wang2019ECANetEC}, SRM \cite{Lee2019SRMAS} and GSop \cite{Wu2018GroupN}, all of which achieved successful records. Recently, transformer \cite{Dosovitskiy2020AnII, git} has become more and more popular. For example, VIT \cite{Dosovitskiy2020AnII} is pioneering the concept of utilizing the transformer encoder for feature extraction and prediction. 
\section{Method}
\subsection{Overview}

For image object detection, the detector is inefficient in detecting small objects since there is a misalignment between different levels in the FPN and information loss due to multiple convolutions, inherent in the fact that the information is uncertain after the feature information has been extracted.
\par For that, this paper proposes a novel Multi to Single (M2S) to improve detector performance for small objects. As shown in Fig.\ref{fig:1}, the M2S is comprised of two modules: Cross-scale Aggregation Module and Dual Relationship Module. 
\par M2S approaches the problem in two ways: firstly, collecting sufficiently rich semantic information, and secondly, refining the information collected. M2S implements the operation in two steps: In the first step, five-level feature from the backbone network are aggregated into three-level feature, and in the second, three-level feature with different characteristics are used in order to enhance the low level of the neck network. Different from PANet, Bi-FPN which creates multiple parallel paths, the M2S proposed in this paper aggregates multiple parallel paths into a single pathway. The five-level feature of the backbone network are fed into the Cross-scale Aggregation Module (CAM) for collecting the semantics of features at different scales and aggregating them.
\par Efficient feature extraction is the first step in M2S, the second part is to enhance the filtering of valid information. In order to refine the aggregated information more efficiently, we have introduced Dual Relationship Module (DRM). Enhanced head network features using DRM to achieve better detection performance. The next paper describes CAM and DRM in detail.
\par  Overall, this paper improves the feature extraction capability of the detector utilizing CAM as well as synthesizes the information from the multi-scale features to provide a foundation for the subsequent work. After obtaining richer feature information the information is fused and weighted into the head network using DRM. Incorporating the above two steps, the improved performance of the detector is obtained.

\subsection{ Cross-scale Feature Aggregation}
Generating three-level feature with rich contextual information is the ambition of CAM. The Cross-scale Fusion Node (CFN) is a sub-module of CAM and is applied to the fusion of adjacent three-layer features. The top-down and then bottom-up workflow, CAM modules are shown in Fig.\ref{fig:1}, in a "V" shape module layout. Where the intermediate layer of input to the CFN is the output of the previous CFN. It will better integrate the bottom-up features of the backbone network while allowing both top and bottom information to interact.
 \begin{figure}[t]
    \centering
    \includegraphics[width=10cm]{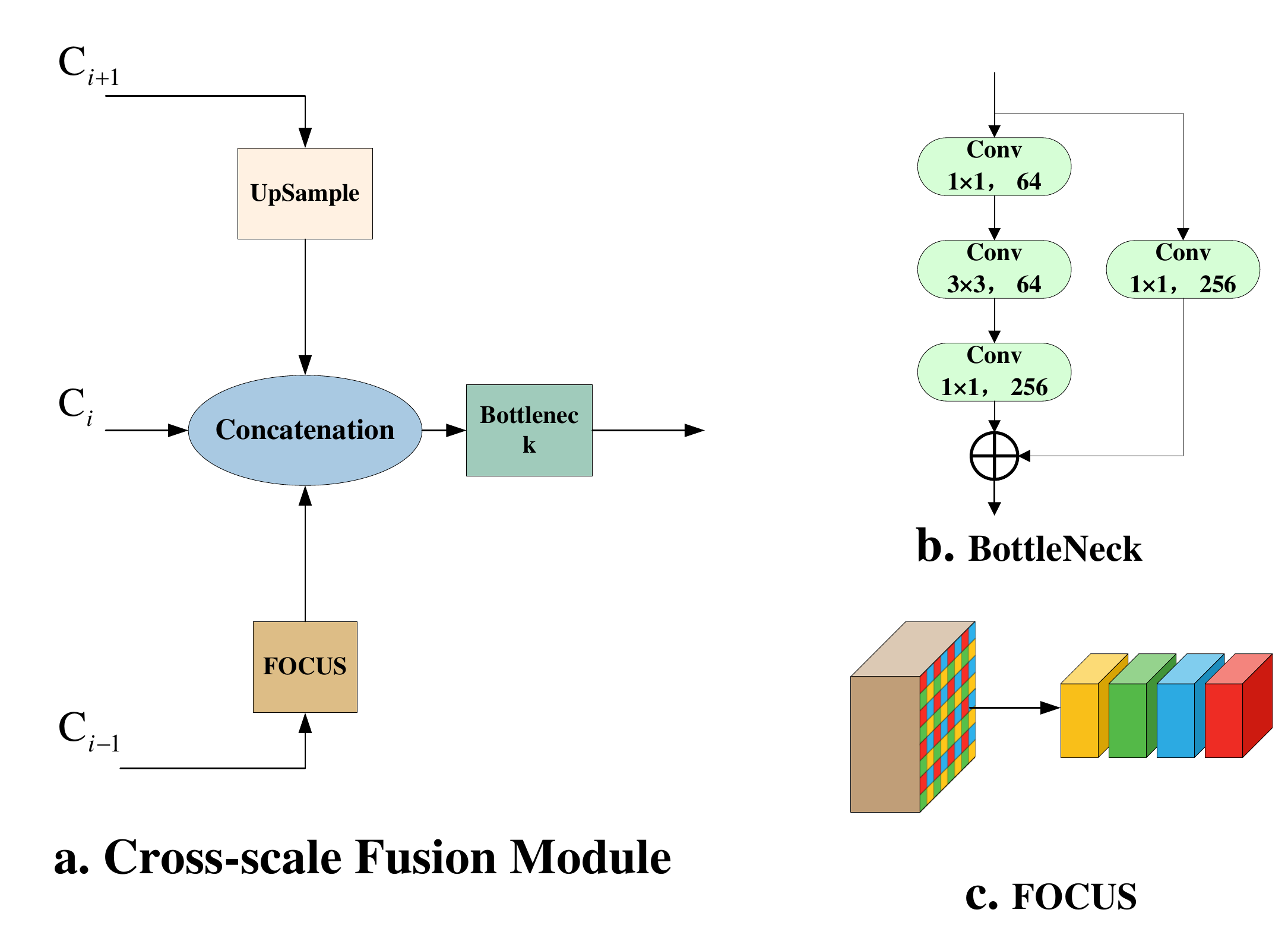}
    \caption{The architecture of CFN, contains two main blocks, the FOCUS block, and the BottleNeck block. a) The three features of the CFN input: low, medium, and high after fusion output a feature. b) BottleNeck as a module for extracting fusion features.  c) FOCUS generates features with four times the number of input channels.}
    \label{fig:2}
\end{figure}

\par The structure of Cross-scale Fusion Module is shown in Fig.\ref{fig:2}. The three adjacent features $ C_{i-1}$ ,$C_{i}$ , $C_{i+1}$ , $(2\leq $i$ \leq 4)$  from the backbone are fed in respectively to the CFM as inputs. It is necessary to pre-process  $C_{i-1}$  and 
 ${\ C}_{i+1}$  because the three features of the input come from levels of different depths. We note that FOCUS down-sampling is done with pixels and channels by reshaping them. We believe that the FCOUs down-sampling approach allows for channel and spatial information interaction. Thus, FCOUs are applied instead of convolution with stride 2 towards $C_{i-1}$   down-sampling. The bilinear interpolation method was used to up-sample  ${\ C}_{i+1}$ . $C_{i-1} $ and  ${\ C}_{i+1}$ are reshape to the same shape as  $ C_i$ . The former two features are concatenated with  $ C_i$  and fed to a Bottleneck module. 
 \par The CAM is made up of several CFNs placed together and the goal of collecting semantics is accomplished through an internal workflow. Eventually, CAM will converge the five-level feature into three feature mapping: Low, Mid, and high.
\subsection{Dual Relationship Module}
The object detection task addresses the questions of "where" and "what". Channel attention and spatial attention mechanisms, which are widely used in computer vision research, are applied to improve the "what" and "where" capabilities of models respectively. Channel attention focuses on "what" is the meaningful input image, while spatial attention focuses on "where" the most informative part is. DRM plays a pivotal role as a bridge between CAM and head network. DRM captures the multidimensional relationship from CAM to enhance and calibrate the input. Each of High, Mid and Low is fed into a module relative to the DRM to obtain richer contextual information.

\begin{figure}[t]
    \centering
    \includegraphics[width=12cm]{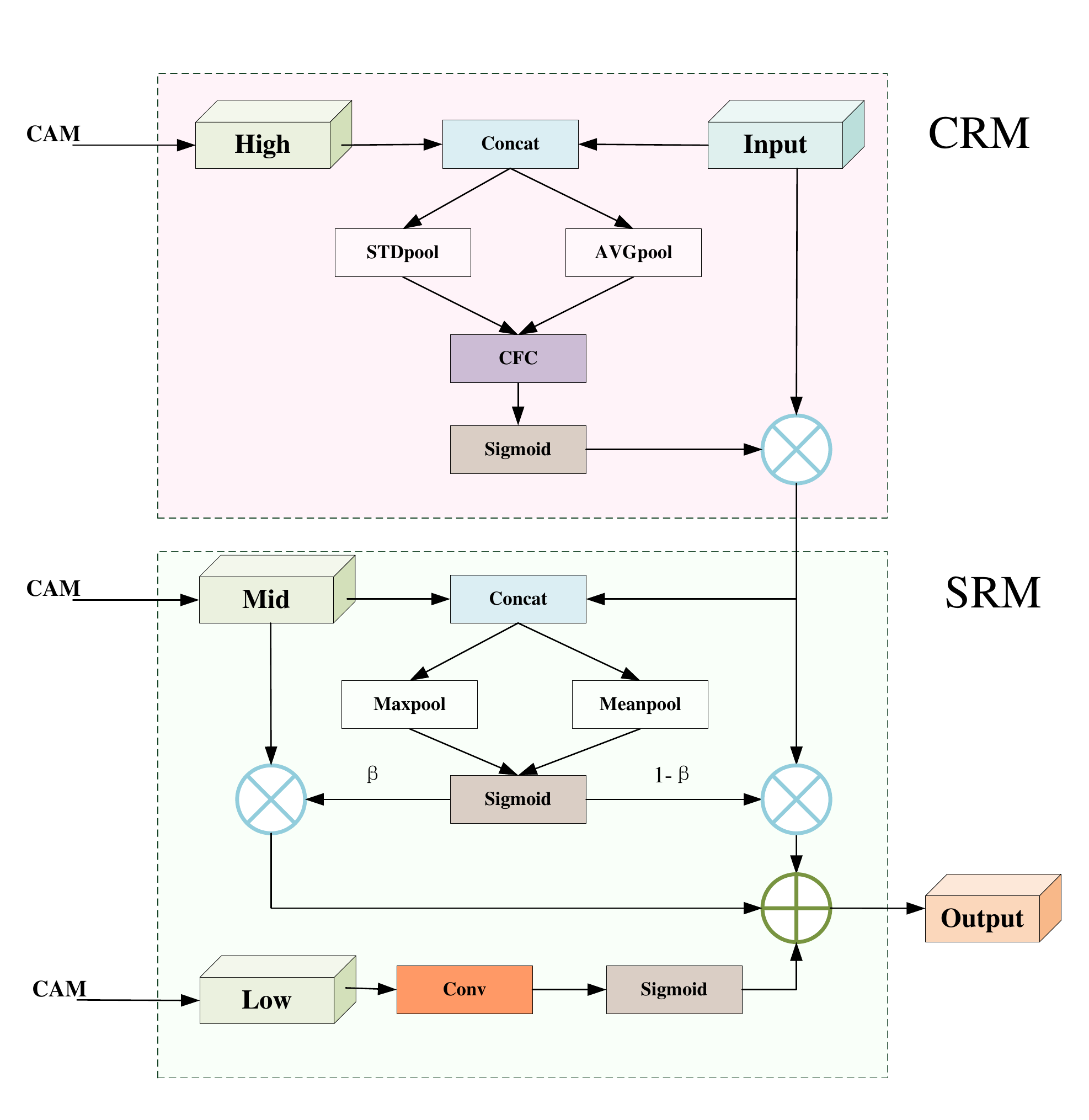}
    \caption{Details of the module. a) The structure of the SRM is shown in the frame with a pink background. b) The structure of the SRM is shown in the green background and the DRM output is also the output of the SRM. c) $\oplus$ and $\otimes$ represent the element-wise sum and matrix multiplication respectively}
    \label{fig:3}
\end{figure}
\subsubsection{Channel Relationship Module}
Inspired by the idea of SRM, ECA et al. embedded channel attention in detectors. The channel attention mechanism enables the efficient extraction of object information. The paper proposes a CRM channel attention module, which is also embedded in the FPN to achieve enhanced detector performance.
\par The details of the CRM are shown in Fig.\ref{fig:3}. The "High" feature map is the Cross-scale Aggregation Module (CAM) output fusion feature. Therefore, we considered it more effective to reinforce input in the channel dimension utilizing "High". Following the concatenation of "High" and input, a convolution is first applied so that the number of channels after the convolution output is the same as input. Next, collects global information by adopting style pooling which combines global average pooling and global standard deviation pooling. Two different style pooling, each able to capture different qualities of different features. Finally, two style pooling obtain the input channel feature information and through a full-connected layer, the layer generates a channel weight vector. Then the parameter of the activation operation is m:

\begin{equation}
m=S_i\left(P\left(A_c\right)\right),
\end{equation}

\par Where S is the Sigmoid activation function. STD and AVG indicate standard pooling and global average pooling respectively. $A_c$ is a feature which concatenate "High" and input$\{A_c, High, input\}\in R^{C\times H\times W}$ The Sigmoid function is adopted to calculate the attention space feature map S:
\begin{equation}
S_i=\ \frac{1}{1\ +\ e^{-z_i}},
\end{equation}
\par Where S calculates the degree of influence of the ith position. Feature information for a given dimension can enrich the information of the feature mapping based on the input. The Sigmoid activation functions in this paper are all formulated as above. Where $z_i\in R^C$ has different representations depending on the type of pooling, e.g. for extracting channel information $z_i$ means the information of the $i -th$ channel.

\par The number of channels in $A_c$ is kept the same as input.Finally, multiply the above results with input to obtain the final output CR as follows. Overall, an SRM can be written as:
\begin{equation}
CR=mB_i ,
\end{equation}
\par Where B is denoted as Input shown in fig.\ref{fig:3}.Thus, through the structure described above, channel attention can selectively focus on important features and suppress unnecessary ones. 
\subsubsection{Spatial Relationship Module}
\par The limitations of channel attention mean that it can only be enhanced and weakened channel-wise. For its part, Spatial Relationship Module (SRM) augments CR with spatial dimensional information. The CR was obtained from the CRM. The details of SRM are shown in Fig.\ref{fig:3}.

\par The "Mid" from a deeper layer also passes more convolutions. The "Low" from CAM contains three shallow levels of feature information. The former contains semantically rich information, while the latter contains accurate structural information. Based on the difference in characteristics between the above two, SRM is also divided into two branches for feature enhancement.
\par First, CR and "Mid" are pooled by a splice then two different types of pooling to obtain spatial information features like CRM. We adopted the Sigmoid activation function to obtain a spatial weight vector. The spatial weight vector will then be multiplied by the CR and "Mid" weighted by a defined parameter $\beta$. Note that $\beta$ is a trainable parameter here, and the initial setting of $\beta$ is 0.3. This branch's end sums the weighted "Mid" and the weighted CR element-wise. After this branch results in E that is weighted by the spatial weights vector, as shown in Eq.3:

\begin{equation}
E=w(\beta\bullet M+(1-\beta)\bullet CR)\ ,
\end{equation}
\par Where the "Mid" feature is denoted as the second layer feature from the CAM. CR is the channel enhancement feature represented in Eq.3. $\beta$ is defined as the assignment weight and is initially set to 0.3. 
\par The second branch is relatively simple compared to the first. The second branch uses the accurate information from "Low" to perform information offset correction for the information from E. A vector with a channel number of 1 is generated after a convolution, with the same width and height as "Low". The bias feature for this branch prediction is then obtained by a Sigmoid activation function. In the end, LF is obtained by summing the bias features with E to obtain the spatial correlation enhancement and alignment, which is expressed by the following equation:
\begin{equation}
LF\ =\ E+\ S(Conv(L)) ,
\end{equation}
\par Where Conv denotes a convolution with a kernel size of $\times1$. E is the feature in eq.3. L denotes the lower feature of the three-level feature map from CAM.
\begin{figure}
    \centering
    \includegraphics[width=5cm]{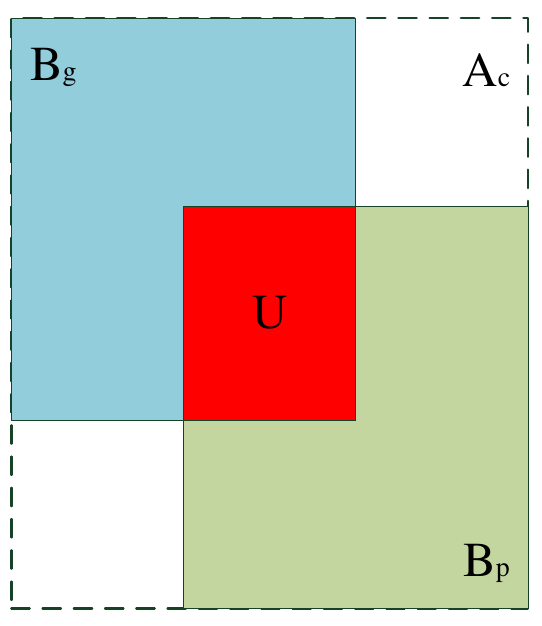}
    \caption{The Intersection over the Union of the ground truth bounding box and the prediction box.}
    \label{fig:6}
\end{figure}
\subsection{Loss function}
Intersection over Union (IoU) is very essential as a basis for judging the accuracy of the size and location of the prediction box. It directly shows the degree of intersection of the ground truth bounding box and the prediction box. The calculation formula of the IoU loss function is as follows:
\begin{equation}
IOU\ =\frac{\ B_g\cap B_p}{B_g\cup B_p} ,
\end{equation}

\par Where the ground truth bounding box is denoted as $B_g$, and the prediction box is denoted as $B_p$. The IoU would not reflect the true quality of the prediction boxes, with the same comparative area but different comparative situations. GIoU introduces a penalty term to better reflect the quality of the prediction box. The calculation formula of the GIoU loss function is as follows:
\begin{equation}
L_{GIoU}\ =1-\ GIoU ,
\end{equation}
\par Where the calculation formula of GIoU is as:
\begin{equation}
GIoU=IoU\ -\ \frac{A_c-U}{A_c} ,
\end{equation}
\par GIoU introduces a new box $A_c$, unlike IoU, which directly calculates the intersection ratio of the two boxes $B_g$ and $B_p$. A is a minimum rectangle that encloses both  $B_g$ and $B_p$. And the combined area of $B_g$ and $B_p$ is U. The IoU loss function, once there is no intersection, the intersection ratio is 0, and optimization cannot be continued. GIoU also provides optimization for loss from the introduced penalty in terms of information on  $B_g$ and $B_p$ when the intersection rate is 0. Therefore, GIoU offers a more precise measure of the intersection ratio than IoU. Objects with smaller areas are better optimized because the GIoU approaches -1 as $B_g$ $\cup$ $B_p$ tend to 0 and are farther apart. GIoU is applied in this detector.

\section{Experiments}

\begin{figure}[t]
    \centering
    \includegraphics[width=12cm]{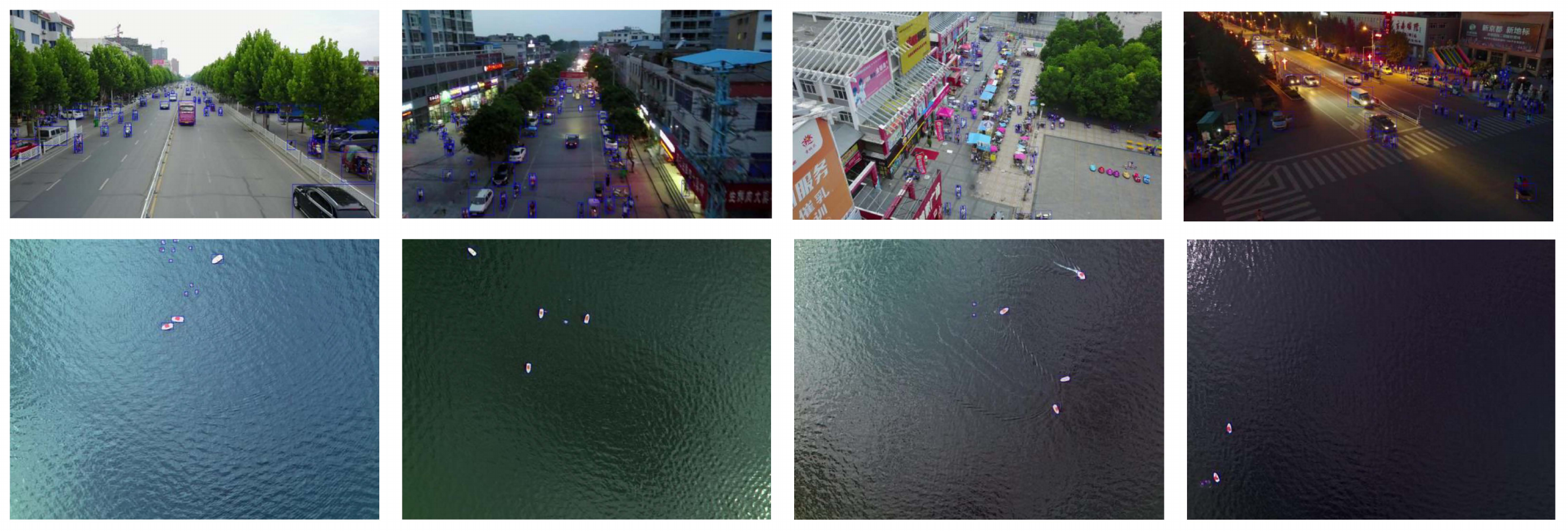}
    \caption{The VisDrone2021-DET~ \cite{Cao2021VisDroneDET2021TV} and SeaDroneSeeV2~ \cite{Varga2021SeaDronesSeeAM} datasets display images. a) The first row of images is the validation set for the VisDrone2021-DET Dataset. VisDrone2021-DET faces a great diversity of challenges such as small object inference, background clutter, and wide viewpoint. b) The second row of images is the validation set for the SeaDroneSeeV2 dataset. SeaDroneSeeV2 train training images with objects distributed over the sea and a background of marine surface, where the objects are frequently masked by waves.}
    \label{fig:4}
\end{figure}

This section will evaluate our approach to the VisDrone2021-DET dataset and SeaDronesSeeV2. Similar to the last three years, VisDrone2021-DET consists of 6, 471 training images, 548 for verification images, and 1,610 for test images, respectively. SeaDronesSeeV2 consists of the training set and the validation set images are 8,930 and 1,547, respectively. VisDrone2021-DET and SeaDronesSeeV2 are both public datasets for small object detection tasks. Therefore, these two datasets were chosen to evaluate the performance of the method. Fig.\ref{fig:4} displays the image of the validation set for both datasets. For comparison with other state-of-the-art methods, this section showed that mean Average Precision (mAP) on testing split.
\par This section includes four parts: Implementation details; Comparisons with the State-of-the-art; Study of ablation; Analysis of other attention modules.
\subsection{Implementation details}
For the experiment, we adopted CSPDarkNet53 as the backbone network for YOLOv5. The entire network is trained on 2 GPUs(NVIDIA GTX-3090) and a CPU (Intel Xeon Gold 6248R). For the network training work, we adopt the AdmW optimizer with its initial learning rate set to $3{\times10}^{-3}$ to train 100 epochs, then next the SGD \cite{Bottou2010LargeScaleML} optimizer with learning rate of $3{\times10}^{-4}$ is applied to train 100 epochs. The momentum parameter is 0.912 and the weight decay is set to $5{\times10}^{-4}$. We implemented the Pytorch framework to construct M2S and gradually added sub-modules to the model for training. Upsampling is achieved by bilinear interpolation and the GIOU \cite{Rezatofighi2019GeneralizedIO} is applied to evaluate the quality of target detection frames.

\subsection{Comparisons with the State-of-the-art}
\begin{table}[ht]
\caption{ Performance comparison of M2S and other algorithms on VisDrone2021-DET testing.\label{tab1}}
\centering
\begin{tabular}{lllllll}
\toprule
Method & AP	& AP50 & AP75 & AR1 & AR10 & AR100\\ 
\midrule
    DetNet59 \cite{Li2018DetNetAB} 	& 	15.26	& 29.23     & 14.34        & 0.26  & 2.57  & 20.87\\
    RefineDet \cite{Zhang2017SingleShotRN} 	& 14.90	    & 28.76	    & 14.08	       & 0.24	& 2.41	& 18.13\\
    RetinaNet \cite{Li2018DetNetAB} 	& 	11.81	& 21.37	    & 14.08	       & 0.21	 & 1.21	 & 5.31\\
    Cascade-RCNN \cite{Cai2017CascadeRD} 	& 	16.09	& 31.91	& 15.01	 & 0.28	 & 2.79	 & 21.37\\
    YOLOv5 \cite{Liu2022SFYOLOv5AL}   	&   15.00	& 28.60	 & 14.20   & - & - & -\\
    Ours    	&   16.10 	& 29.70 & 15.19    & 0.29 & 2.90 & 22.10\\
\bottomrule
\end{tabular}
\end{table}
\unskip
Compare the experimental results of the CAM and DRM with state-of-the-art (SOTA) object detection methods on the VisDrone2022-DET testing set in Table 1. The comparison of multiple types of detectors, including one-stage, two-stage, and multi-stage detectors and different backbone networks. Each SOTA result and M2S result in input size has the same input size (input size =640X640) as the YOLOv5s default. It is worth noting that the M2S achieves AP (at IoU=.50:.05:.95) of 16.10\%, which has a similar performance with the two-stage detector Cascade-RCNN. The performance of the one-stage detector is superior to that of previous the one-stage detectors e.g., RefineDet, RetinaNet, and DetNet59 in terms of AP. The method proposed in this paper has not only improved the AP but also the AR (average recall rate) in comparison to the baseline. Among one-t detectors, "Yolov5s+ours" exceeds its AP metric by 0.84\% relative to the higher-performing DetNet59, although Yolov5s is inferior to DetNet59. With the M2S module boost, the Yolov5s performance can overtake the one-stage detectors in the Table \ref{tab1} on the contrary, and can even better the Cascade-RCNN.
\begin{figure}[t]
    \centering
    \includegraphics[width=14cm]{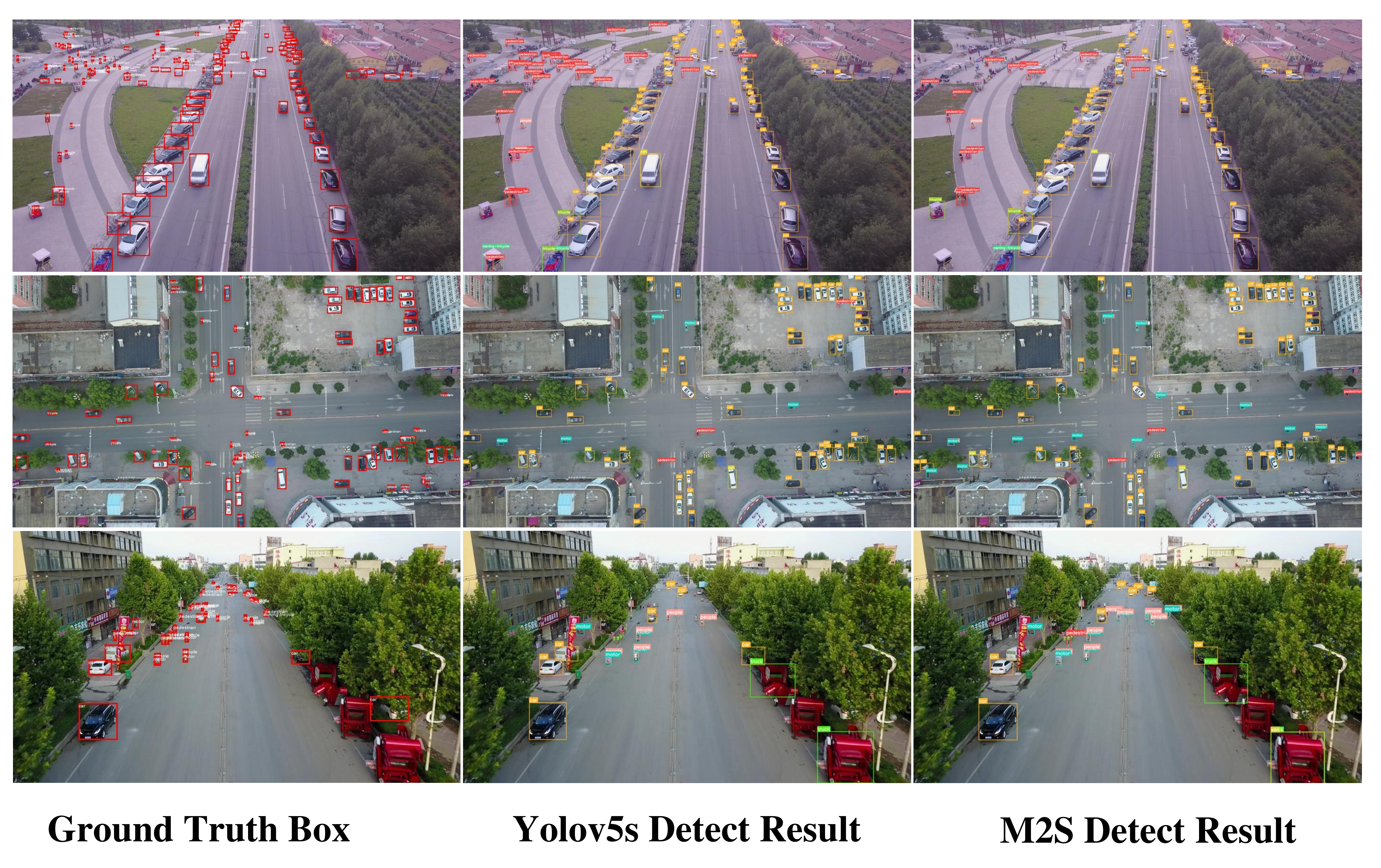}
    \caption{Detection effect of M2S on VisDrone2021-DET. a) Ground truth of VisDrone2021-DET val set. b) Predicted results of Yolov5s on val set. c) Predicted results of M2S on val set.}
    \label{fig:5}
\end{figure}

\begin{figure}[t]
    \centering
    \includegraphics[width=14cm]{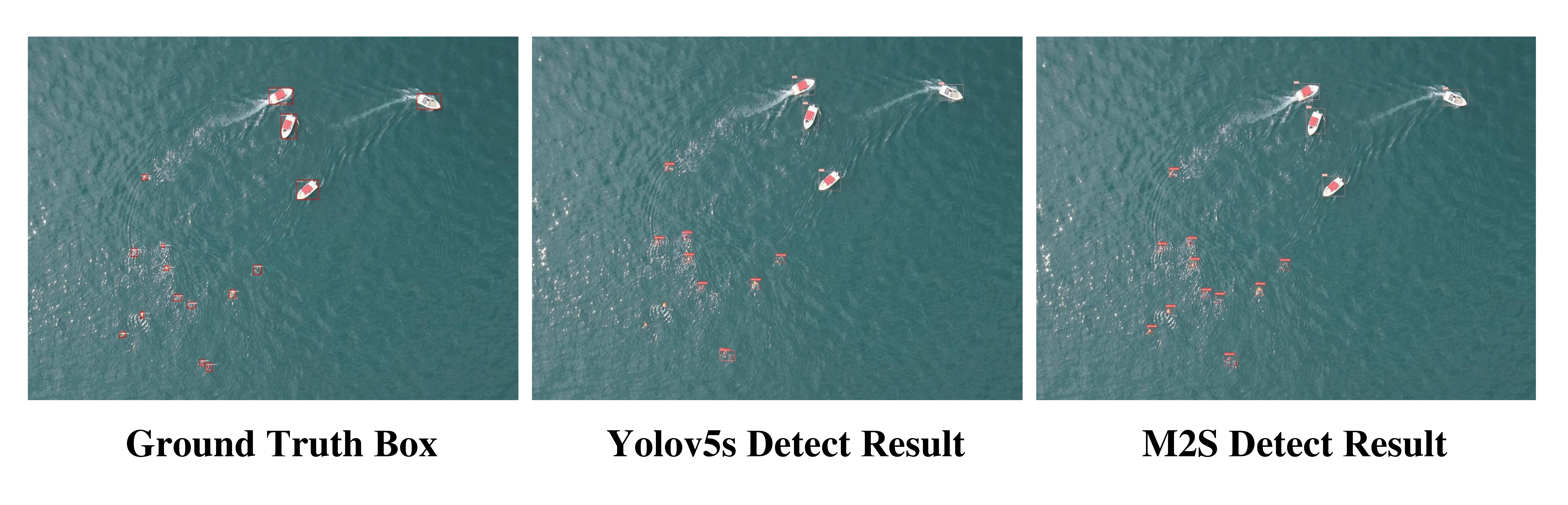}
    \caption{Detection effect of M2S on SeaDronesSeeV2. For better display of the results, the display image has been enlarged and cropped. a) Ground truth of SeaDronesSeeV2 val set. b) Predicted results of Yolov5s on val set. c) Predicted results of M2S on val set.}
    \label{fig:7}
\end{figure}
\par Besides VisDrone2021-DET this section also validates the performance of our method on SeaDronesSeeV2. Training using transfer learning to SeaDronesSeeV2. The same training strategy as on VisDrone2021-DET was used to train the SeaDronesSeeV2 in this experiment. The input image size is also set to 640x640. The results of the performance on the SeaDronesSeeV2 val set compared to SOTA are shown in Table \ref{tab2}. As the Table \ref{tab2} shows, our method outperforms the baseline, with a 15.68\% improvement in the AP metric. YOLOv5s is 12.37\% lower than YOLOv7 in the AP metric. However, YOLOv5s overtakes YOLOv7 by 3.31\% after combining our method. In addition, the combination of the M2S module increases the performance of the Yolov5s significantly, achieving more than 26.65\% of the AP50 index, which was similar to the performance of the Faster R-CNN ResNet-18 Baseline.

\begin{table}
 \caption{ Performance comparison of M2S and other algorithms on SeaDronesSeeV2 val set.}
  \centering
  \begin{tabular}{llllll}
    \toprule
    Method & AP	& AP50 & AP75 & AR1 & AR10 \\ 
    \midrule
    YOLOv7 \cite{Wang2022YOLOv7TB} 	& 	42.00	& 72.00     & 42.00        & 36.00  & 49.00 \\
    FRCNN-RN \cite{Girshick2015FastR} 	& 24.34	    & 52.17	    & 20.4	       & 23.58	& 32.15	\\
    YOLOv5 \cite{Liu2022SFYOLOv5AL} & 	29.63	& 52.40	    & 28.46	       & -	 & -	 \\
    Ours 	& 	45.31 	& 78.82 	& 43.50 & 36.50	 & 46.9	 \\
\bottomrule
  \end{tabular}
  \label{tab2}
\end{table}
\par Combined with the results of the two experiments in this section, it appears that our method applied to YOLOv5s can improve the performance of the detector. However, a point worth noting is that our method performs considerably inferiorly on the VisDrone2021-DET compared to the SeaDronesSeeV2. The images in VisDrone2021-DET are made to shift the size of the objects in the picture because of the altitude varying of the drone flight. And most of the images in SeaDronesSeeV2 were obtained by keeping the drone at the same height. Combining the characteristics of the two datasets, it may be that the scale of VisDrone2021-DET which object varies dramatically, causing the objects to exceed the range of scales enhanced by our method. Our approach is therefore more suitable for tasks designed with small objects to improve the performance of the detector.

\subsection{Ablation Studies}
\subsubsection{Study of ablation}
CSPDarkNet53 is the backbone for this section, and the methods we propose will all be compared with CSPDarkNet53. The method we propose is validated by the comparison of several experimental groups. The results of the ablation experiments are shown in Table \ref{tab3}. In the experimental design, CAM and DRM are embedded in CSPDarkNet53 respectively. An improvement of 0.57\% and 0.73\% points in the AP metric was achieved in CAM and DRM embedded into the detector respectively. It is worth noting that although the two modules were embedded to achieve a 2.08\% improvement, the improvement achieved with one module alone is not satisfactory. This phenomenon occurs, and presumably it is difficult to extract and use the information effectively, in the case of a single module embedded separately. Merely enhancing ability to extract information or improving refinement will not achieve the best results. This is why the combination of CFM and DRM is necessary and inseparable.

\subsubsection{Visual analysis of M2S}
\par Our method detect results in this section are shown in Fig.\ref{fig:5} and Fig.\ref{fig:7}. Comparing baseline on VisDrone2021-DET with a confidence threshold is 0.25, our method can identify more objects of very small size. And on SeaDronesSeeV2 with a confidence threshold is 0.25, our method can identify swimmers that are small in size and obscured by waves. Combining Fig.\ref{fig:5} and Fig.\ref{fig:7}, it is intuitively clear that for smaller-sized object detection, our method performs even better. But for larger-sized object detection, our method has a similar performance to Yolov5s. Accordingly, detection performance for small objects is a major enhancement within the M2S.
\begin{table}
\caption{Ablation study of M2S on VisDrone2021-DET val and 640×640 resolution as input.\label{tab3}}
\centering
\begin{tabular}{lllll}
\toprule
\textbf{CAM}	& \textbf{DAM}	& \textbf{AP} & \textbf{AP50} & \textbf{AP75}\\
\midrule
     	    & 			& 17.90        & 33.00       & 17.30\\
$\surd$  	& 			& 18.47 (+0.57)        & 34.37 (+1.37)       & 18.80 (+1.50)\\
        	& $\surd$			& 18.63 (+0.73)        & 35.02 (+2.02)      & 17.32 (+0.02)\\
$\surd$     & $\surd$		& 19.98 (+2.08)        & 36.30 (+3.30)      & 18.50 (+1.20)\\
\bottomrule
\end{tabular}
\end{table}
\unskip
\subsection{Analysis of other attention modules}
To discuss the performance differences between our approach and other attention modules, this section has been designed as an experiment. The detection accuracy is placed in the order as "CSPDarkNet53 +CAM + DRM, CSPDarkNet53 +CAM + CBAM, CSPDarkNet53 +CAM + ECA, CSPDarkNet53 +CAM + SE, CSPDarkNet53 +CAM, CSPDarkNet53 +CAM, and CSPDarkNet53". As all other attention modules are single input and single output, "Mid" is used as an input to ensure that the output is the same size as the next stage feature. As can be seen from the Table \ref{table4}, DRM in combination with CAM performs the best, outperforming CBAM by 0.64\% in the AP index and even by 1.38\% in the SE. Even the ECA has a good performance, slightly inferior to the CBAM more so than our method.
\par According to Table \ref{table4}, it can be seen that modules with a combination of multiple attention mechanisms achieve better performance compared to modules with a single attention mechanism. The CBAM with a combination of multiple attention mechanisms also has performance differences over our approach. Multiple inputs bringing a wealth of information may be the cause of this phenomenon. It is evident that better performance can be achieved with more information combined with dual attention mechanisms than with a single input with dual attention mechanisms. It is also on this basis that our approach is able to obtain a better performance than other attention modules.

\begin{table}[t]
\caption{ Experimental results of YOLOv5s on VisDrone2021-DET for different attention modules.\label{table4}}
\centering
\begin{tabular}{ p{6cm}|p{2cm}|p{2cm}  }
 \hline
Method & AP & AP50  \\ 
 \hline
 CSPDarkNet53 		& 17.90        & 33.00 \\
    CSPDarkNet53 +CAM 		& 18.47 (+0.57)        & 34.37 (+1.37) \\
   
    CSPDarkNet53 +CAM + CBAM \cite{Woo2018CBAMCB} 		& 19.34 (+1.44)        & 35.44 (+2.44) \\
    CSPDarkNet53 +CAM + SE \cite{Hu2017SqueezeandExcitationN}		& 18.60 (+0.70)        & 34.22 (+1.22) \\
    CSPDarkNet53 +CAM + ECA \cite{Wang2019ECANetEC}		& 19.12 (+1.22)        & 35.15 (+2.15) \\
 CSPDarkNet53 +CAM + DRM 		& 19.98 (+2.08)    & 36.30 (+3.30) \\
 \hline
\end{tabular}
\end{table}

\section{Conclusions}

This paper introduces a multi-to-single module. On the one hand, the method connects multi-level backbone network features across scales as multiple inputs, efficiently extracting and aggregating features at multiple scales. On the other hand, the channel attention mechanism and the spatial attention mechanism, both dimensions of attention, are used to refine and filter the information. Ultimately, the high-resolution features in the head network are enhanced using our method to achieve improved detector performance. Experiments on VisDrone2021-DET and SeaDronesSeeV2 datasets demonstrates that the Cross-scale Aggregation Module and Dual Relationship Module can achieve satisfactory improvements.

\bibliographystyle{unsrt}  

\bibliography{references.bib}

\end{document}